\documentclass{article}
\usepackage[preprint]{neurips_2024}

\usepackage[utf8]{inputenc}
\usepackage[T1]{fontenc}
\usepackage{microtype}
\usepackage{needspace}

\usepackage{amsmath}
\usepackage{amssymb}
\usepackage{booktabs}
\usepackage{array}
\usepackage{graphicx}
\usepackage{fontawesome5}
\usepackage{enumitem}
\usepackage{xcolor}

\usepackage{url}
\usepackage[hidelinks]{hyperref}
\setcitestyle{numbers,square,comma}
\definecolor{rulegray}{gray}{0.72}   
\definecolor{tagcolor}{gray}{0.42}   
\definecolor{glosslink}{rgb}{0,0,0}  

\providecommand{\textls}[2][]{#2}

\newcommand{\cathead}[1]{%
  \par\addvspace{15pt}\needspace{5\baselineskip}%
  \noindent{\normalsize\scshape\bfseries\textls[70]{#1}}\par
  \vspace{2.5pt}%
  {\color{rulegray}\hrule height 0.6pt}%
  \par\addvspace{8pt}%
}

\DeclareRobustCommand{\glref}[2]{\hyperlink{gl:#1}{\textcolor{glosslink}{#2}}}
\newcommand{\glentry}[2]{\item[\hypertarget{gl:#1}{#2}]}
\newenvironment{glosslist}
  {\begin{description}[style=nextline,leftmargin=2.1em,labelindent=0pt,
      itemsep=5pt,parsep=0pt,topsep=2pt,font=\normalfont\bfseries]}
  {\end{description}}

\newcounter{faqnum}
\renewcommand{\thefaqnum}{Q\arabic{faqnum}}
\newcommand{\faqq}[1]{%
  \par\addvspace{10pt}\needspace{6\baselineskip}%
  \refstepcounter{faqnum}%
  \noindent\hangindent=2.1em\hangafter=1
  \makebox[2.1em][l]{\color{tagcolor}\bfseries\small\thefaqnum}%
  \textbf{#1}\par\vspace{2pt}%
}
\newenvironment{faqa}
  {\par\begingroup\leftskip=2.1em\noindent\ignorespaces}
  {\par\endgroup\addvspace{1pt}}

\DeclareRobustCommand{\wer}{\glref{wer}{WER}}
\DeclareRobustCommand{\per}{\glref{per}{PER}}
\DeclareRobustCommand{\cer}{\glref{cer}{CER}}

\newcommand{\yes}{\checkmark}
\newcommand{\no}{--}

\title{Myovox: Reading Speech from the Muscles of the Face}

\author{%
  Varshith Madishetty\\
  \texttt{madishettyvarshith@gmail.com}%
}

\begin{document}

\maketitle

\vspace{-6pt}
\begin{center}
\small
\faGithub~\href{https://github.com/Varshith-0/myovox}{\texttt{github.com/Varshith-0/myovox}}\qquad
\faGlobe~\href{https://varshith-0.github.io/myovox/}{\texttt{varshith-0.github.io/myovox}}
\end{center}
\vspace{2pt}
{\color{rulegray}\hrule height 0.6pt}
\vspace{6pt}

\begin{abstract}
Myovox, from \emph{myo} (muscle) and \emph{vox} (voice), decodes
\glref{openvocab}{open-vocabulary} English text from 31-channel
\glref{semg}{surface electromyography (sEMG)} recorded from the muscles of the face during
\glref{vocalized}{vocalized speech}. It takes the single-subject \emph{emg2speech}
\glref{corpus}{General Corpus} from a published 51.17\% word error rate to 18.53\%, in three
separable moves, each measured in isolation.
First, I recover the open-vocabulary decode settings missing from the public release and
reach a faithful \textbf{40.63\% \wer{} / 39.02\% \per{}} baseline whose phone error rate matches
the published one to within 0.8 points, so the \glref{am}{acoustic model} is reproduced faithfully.
Second, I replace the \glref{causal}{causal} encoder with a \glref{bidir}{bidirectional}
\glref{conformer}{Conformer} trained by a four-term \glref{crossmodal}{cross-modal distillation}
against the parallel audio's \glref{wavlm}{WavLM-Large} layer-9 features, reaching \textbf{26.14\%
\wer{} / 22.34\% \per{}} from the \glref{emg}{electromyography} alone. Third, I
\glref{ensemble}{ensemble} two acoustic models, union their multi-scale \glref{nbest}{$n$-best}
lists, and \glref{rerank}{rerank} with a \glref{qlora}{QLoRA}-fine-tuned 7B
\glref{lm}{language model}, reaching \textbf{18.53\% \wer{}}, the best result reported on this
corpus, though not the best reported for sEMG-to-text on other corpora (Section~\ref{sec:related}).
I then report the negative result that bounds the whole approach: reranking is exhausted at 18.5\%
because the binding constraint is the electromyographic \emph{acoustic} phone error rate
($\sim$20.9\%), not the language model. The correct words are simply absent from the acoustic
posteriors, so no reranker can reach the 9.30\% $n$-best \glref{oracle}{oracle}. All test numbers
are on the 400-sentence held-out \glref{test}{test set} under the authors' official
8{,}500\,/\,760\,/\,400 \glref{split}{sequential split}; every hyperparameter is tuned once on
\glref{val}{validation} and applied once to test.
\end{abstract}

\section{Introduction}

When you speak, the decision to say a word reaches the muscles of your face, jaw, and throat long
before any sound leaves your mouth. Those muscles fire, and the firing shows up as faint electrical
voltages on the skin. Myovox reads that electricity. A grid of 31 sensors records the voltages a few
thousand times a second, and a chain of models turns the recording into a line of English text. The
name is literal: \emph{myo} for the muscle, \emph{vox} for the voice the muscle was reaching for.

The long-term goal of \glref{emg}{electromyographic} speech decoding is to help people who cannot
speak because of conditions such as ALS or a laryngectomy. This report does not demonstrate that.
Myovox was trained entirely on a healthy speaker who vocalized normally, so I cannot claim that it
works for people who have lost their voice. What I can say is that it demonstrates a different use
case: allowing a healthy person to communicate without producing audible speech, where the intention
to speak is decoded from facial muscle activity instead of sound. I view this as a stepping stone
toward true \glref{silent}{silent-speech} systems rather than the destination.

The report is organized around a single discipline: start from a strong published result, and measure
the effect of each change in isolation before applying the next. That is what makes the final number
decomposable into a decode fix, a modeling gain, and an extraction gain, rather than a single
undifferentiated delta. The starting point is the \emph{emg2speech} work of Gowda et
al.~\citep{gowda2026emg2speech}, whose Appendix~D.4 (``emg2text'') reports 51.17\% \wer{} and 38.19\%
\per{} on the General Corpus. The three sections that follow trace three moves: a decode correction
that recovers most of the published gap while matching the original phone error rate
(Section~\ref{sec:baseline}); a full-context acoustic model trained to imitate the parallel audio
(Section~\ref{sec:acoustic}); and a decode-time stack of ensembling, $n$-best union, and
language-model reranking (Section~\ref{sec:final}). Table~\ref{tab:overview} is the whole story in
one place. Section~\ref{sec:limit} carries the report's main finding: why the system stops improving
at 18.5\%, and where the next gain has to come from. Terms
of art are collected in the Glossary (Section~\ref{sec:glossary}), and every occurrence of a glossary
term in the text links to its entry; the objections the project actually drew are answered in
Section~\ref{sec:faq}.

Two numbers run through everything. \emph{\glref{wer}{Word error rate}} (\wer{}) is what a reader
cares about: the fraction of words wrong after substitutions, insertions, and deletions.
\emph{\glref{per}{Phone error rate}} (\per{}) is the honest gauge of the acoustic model: the fraction
of \glref{phoneme}{phonemes} wrong, measured by \glref{greedy}{greedy decoding} with no
\glref{lexicon}{lexicon} and no language model. \per{} cannot be improved by a better decoder or a
better language model. As it turns out, \per{} is what limits the entire system.

\begin{table}[t]
\centering
\small
\caption{The whole report in one table. All test numbers are on the 400-sentence held-out set under
the authors' 8{,}500\,/\,760\,/\,400 sequential split. \per{} is the decoder-independent greedy CTC
phone error rate. Bold marks each stage's headline word error rate.}
\label{tab:overview}
\setlength{\tabcolsep}{5pt}
\begin{tabular}{@{}c l cc cc@{}}
\toprule
 & & \multicolumn{2}{c}{Validation} & \multicolumn{2}{c}{Test} \\
\cmidrule(lr){3-4}\cmidrule(lr){5-6}
 & System & \wer{} & \per{} & \wer{} & \per{} \\
\midrule
 & Gowda et al., Appendix~D.4 (target)~\citep{gowda2026emg2speech}
   & n/a & n/a & 51.17 & 38.19 \\
\addlinespace[2pt]
1 & Causal TDS + dual-CTC, \emph{corrected} decode
   & 53.12 & 45.31 & \textbf{40.63} & 39.02 \\
2 & Bidirectional Conformer + WavLM-L9 distillation
   & 35.54 & 27.47 & \textbf{26.14} & 22.34 \\
3 & Ensemble $\rightarrow$ $n$-best union $\rightarrow$ LIFT rerank
   & n/a & n/a & \textbf{18.53} & 20.90$^{\dagger}$ \\
\bottomrule
\end{tabular}

\vspace{3pt}
{\footnotesize $^{\dagger}$Phone error rate of the acoustic \glref{ensemble}{ensemble} beneath the
pipeline. Reranking operates on words and cannot change \per{}, which is precisely the point of
Section~\ref{sec:limit}.\par}
\end{table}

\section{Related work}
\label{sec:related}

Two non-invasive routes lead from a person's intention to speak to a line of text, and they differ in
where along the motor chain they tap. \emph{Surface electromyography} reads the muscles of the face
and neck as they articulate, which is the last link before sound. \emph{Electro- and
magnetoencephalography} read cortex, which is many links earlier. Myovox belongs to the first family.
Table~\ref{tab:related} places it against both. Invasive intracortical decoders are excluded
throughout: they are the performance ceiling, but they are not an alternative Myovox competes with.

\paragraph{Surface EMG to text: the Gaddy line.}
The reference corpus for open-vocabulary sEMG speech decoding is the one released by Gaddy and
Klein~\citep{gaddy2020digital}: eight facial channels at 1\,kHz, roughly nineteen hours from a single
English speaker, recorded in both \glref{silent}{silent} and \glref{vocalized}{vocalized} modes with
audio available for the vocalized half. Their first system synthesized speech from silent EMG and was
scored by transcribing that speech; a Transformer front-end and an auxiliary phoneme loss then cut the
open-vocabulary error substantially~\citep{gaddy2021improved}. Gaddy's thesis~\citep{gaddy2022thesis}
moved from synthesis to direct EMG-to-text and set the prior state of the art at 28.8\% \wer{} on
silent EMG and 23.3\% on vocalized EMG. MONA LISA~\citep{benster2024mona} then combined cross-modal
alignment (a contrastive objective binding EMG and audio latents, plus training on audio-only
LibriSpeech) with an LLM reranker, and reached \textbf{12.2\% \wer{} on silent EMG and 3.7\% on
vocalized EMG}. Both halves of that system have direct descendants here: the
\glref{crossmodal}{cross-modal distillation} of Section~\ref{sec:acoustic} and the
\glref{lift}{LIFT}-style reranker of Section~\ref{sec:final}. A recent line asks the harder
question of what is achievable from unvoiced EMG with no paired voiced recordings or audio at
all~\citep{mohapatra2025llms}, which is the setting a patient would actually present.

\paragraph{Myovox is not the best sEMG-to-text system; it is the best one on this corpus.}
This distinction matters enough to state before the table rather than after it. MONA LISA's 3.7\% on
vocalized EMG is 15 points better than the 18.53\% reported here, and it was obtained two years
earlier. The two numbers are not measured on the same recordings and are not directly comparable: the
Gaddy corpus is eight channels and about nineteen hours with both speaking modes from one speaker,
while the \emph{emg2speech} \glref{corpus}{General Corpus} is 31 channels and 9{,}660 sentences from a
different speaker in one mode; MONA additionally trains against an audio-only corpus that Myovox does
not use. What can be said cleanly is narrower: on the \emph{emg2speech} General Corpus, 18.53\% is the
best result I am aware of, against a published 51.17\%. Whether the residual gap to MONA is a property
of the corpus, of the \glref{cov}{covariance front-end}, or of the model is precisely what a single
corpus cannot resolve; \ref{faq:mona} takes the question up directly.

\paragraph{Non-invasive brain to text.}
The parallel effort decodes cortex rather than muscle. Brain2Qwerty~\citep{levy2025brain2qwerty}
decoded typed sentences from 35 volunteers, reaching a 32\% \cer{} from MEG and 67\% from EEG. Its
successor~\citep{zhang2026brain2qwerty2} scaled the data to roughly 22{,}000 sentences across nine
subjects and reached 39\% \wer{} from MEG. Two things about it are worth a reader's attention here.
The first is architectural convergence: independently of this work, that system also settles on a
\glref{conformer}{Conformer} encoder trained with \glref{ctc}{CTC} and a
\glref{lora}{LoRA}-fine-tuned LLM on top: the same two components as Sections~\ref{sec:acoustic}
and~\ref{sec:final}, arrived at from a different signal. The second is that it independently
reproduces the finding of Section~\ref{sec:limit}. Their encoder alone reaches 55\% \wer{}; an
$n$-gram \glref{lm}{language model} takes it to 43\%; the fine-tuned LLM takes it to 39\%. There, the
LLM \emph{worsens} the character-level metric (31\% \cer{} against the encoder's 28\%) even as it
improves the word-level one, and the authors conclude that final performance is dominated by upstream
encoder quality. What transfers to Section~\ref{sec:limit} is that conclusion, not the mechanism, and
the difference is worth stating: their LLM generates the sentence autoregressively and is conditioned
on MEG-derived word embeddings as well as on the encoder's characters, so it can and does degrade the
character metric, whereas the reranker of Section~\ref{sec:final} sits downstream of the greedy CTC
output and cannot move \per{} at all. Their own text-only ablation, which leaves the LLM nothing but
the encoder's characters, is the closer analogue, and it is worth about six \wer{} points over the
encoder alone. Either way a language model bolted onto a weak encoder buys a bounded, single-digit
number of \wer{} points. That is the same wall, found on a different signal.

\begin{table}[t]
\centering
\footnotesize
\caption{Non-invasive speech and language decoding, grouped by corpus. Numbers are as reported by
each work and are \emph{not} comparable across groups: corpora, electrode counts, speaking modes,
subject counts, and metrics all differ. ``Aud.'' marks whether parallel or external audio is used as
a training signal (never at inference). Invasive intracortical systems are omitted by design.}
\label{tab:related}
\setlength{\tabcolsep}{4.5pt}
\begin{tabular}{@{}l l l c ccc@{}}
\toprule
System & Mode & Vocab & Aud. & \wer{}\,$\downarrow$ & \per{}\,$\downarrow$ & \cer{}\,$\downarrow$ \\
\midrule
\multicolumn{7}{@{}l}{\emph{Facial sEMG $\rightarrow$ text: Gaddy corpus} (8~ch, 1\,kHz, $\sim$19\,h, 1 speaker)~\citep{gaddy2020digital}} \\
\addlinespace[1pt]
Gaddy \& Klein 2020~\citep{gaddy2020digital}$^{\ast}$   & silent    & open & \yes & 68.0 & \no & \no \\
Gaddy \& Klein 2021~\citep{gaddy2021improved}$^{\ast}$  & silent    & open & \yes & 42.2 & \no & \no \\
Gaddy 2022~\citep{gaddy2022thesis}                      & silent    & open & \yes & 28.8 & \no & \no \\
Gaddy 2022~\citep{gaddy2022thesis}                      & vocalized & open & \yes & 23.3 & \no & \no \\
MONA LISA~\citep{benster2024mona}                       & silent    & open & \yes & 12.2 & \no & \no \\
MONA LISA~\citep{benster2024mona}                       & vocalized & open & \yes & \textbf{3.7} & \no & \no \\
\addlinespace[3pt]
\multicolumn{7}{@{}l}{\emph{Facial sEMG $\rightarrow$ text: emg2speech General Corpus} (31~ch, 5\,kHz, 9{,}660 sentences, 1 speaker)~\citep{gowda2026emg2speech}} \\
\addlinespace[1pt]
Gowda et al.\ 2026, App.~D.4~\citep{gowda2026emg2speech} & vocalized & open & \yes & 51.17 & 38.19 & \no \\
Myovox, corrected decode (\S\ref{sec:baseline})          & vocalized & open & \yes & 40.63 & 39.02 & \no \\
Myovox, Conformer + distillation (\S\ref{sec:acoustic})  & vocalized & open & \yes & 26.14 & 22.34 & \no \\
Myovox, Conformer, EMG-only (\S\ref{sec:acoustic})       & vocalized & open & \no  & 26.10 & 23.71 & \no \\
\textbf{Myovox, full pipeline} (\S\ref{sec:final})       & vocalized & open & \yes & \textbf{18.53} & 20.90$^{\dagger}$ & \no \\
\addlinespace[3pt]
\multicolumn{7}{@{}l}{\emph{Non-invasive brain $\rightarrow$ text} (typed sentences, healthy volunteers)} \\
\addlinespace[1pt]
Brain2Qwerty v1, EEG~\citep{levy2025brain2qwerty}         & typing & open & \no & \no & \no & 67 \\
Brain2Qwerty v1, MEG~\citep{levy2025brain2qwerty}         & typing & open & \no & \no & \no & 32 \\
Brain2Qwerty v2, MEG~\citep{zhang2026brain2qwerty2}       & typing & open & \no & 39 & \no & 31 \\
\bottomrule
\end{tabular}

\vspace{3pt}
{\footnotesize $^{\ast}$These two systems \emph{synthesize speech} and are scored by transcribing the
synthesized audio, so their \wer{} is an intelligibility measure rather than a direct text-decoding
score. Brain2Qwerty v1 reports only \cer{}; v2 reports both. Brain2Qwerty v1 averages 35 subjects
(19\% \cer{} for the best); v2 averages 9 subjects (22\% \wer{} for the best).\\
$^{\dagger}$The \per{} of the acoustic \glref{ensemble}{ensemble} beneath the full pipeline, not of the
pipeline: reranking operates on words and cannot change \per{}.\par}
\end{table}

\paragraph{What is new here.}
Against that background, this report contributes three things, none of which is a new architecture.
It supplies the open-vocabulary \glref{blank}{decode settings} missing from the \emph{emg2speech}
release, without which that release decodes at roughly 75\% \wer{} rather than the 40.63\% reported
here, and shows by a matched \glref{per}{phone error rate} that the acoustic model is unchanged
(Section~\ref{sec:baseline}). It prints a control that undercuts its own favoured component: an
encoder trained on \glref{emg}{electromyography} alone matches the audio-distilled one at the word
level, so the parallel audio is not a hidden crutch beneath the headline number
(Section~\ref{sec:acoustic}). And it locates, with three independent lines of evidence, the point at
which decode-time machinery stops paying (Section~\ref{sec:limit}).

\section{The data}
\label{sec:data}

The recordings are the healthy-subject \emph{\glref{corpus}{General Corpus}} from the
\emph{emg2speech} release. Its properties are summarized in Table~\ref{tab:data}. Three facts about
it shape the rest of the report.

\paragraph{It is one person.}
The entire system is trained and evaluated on recordings from a single healthy speaker. I therefore
do not know how well it generalizes to other people, anatomies, or speaking styles. The results in
this report should be read as a demonstration of what is possible for one speaker, not as evidence
that the same performance will transfer to others.

\paragraph{It has parallel audio.}
While the subject spoke, a microphone recorded the voice at the same time; per-sentence
electromyography and audio durations correlate at 1.000. That audio is the training signal I lean on
hardest in Section~\ref{sec:acoustic}. It is also a crutch that will not exist at
\glref{silent}{silent-speech} inference, so I treat it as a training-time-only teacher and, at the
end, check how much the system depends on it.

\paragraph{It is open-vocabulary.}
The decoder works over a 34{,}546-word \glref{librispeech}{LibriSpeech}-derived
\glref{lexicon}{lexicon}~\citep{panayotov2015librispeech}, roughly five times the corpus vocabulary,
so the system can emit words it never saw in training. Twelve of the 2{,}429 test tokens fall
\glref{oov}{outside the lexicon}, a 0.49\% \wer{} floor that no amount of modeling can remove.

\begin{table}[t]
\centering
\small
\caption{The \emph{emg2speech} General Corpus.}
\label{tab:data}
\begin{tabular}{@{}l p{8.6cm}@{}}
\toprule
Property & Value \\
\midrule
Sentences & 9{,}660 (9{,}541 unique) \\
Subject & single, healthy \\
Electromyography & 31-channel surface array, 5\,kHz \\
Parallel audio & recorded simultaneously (per-sentence duration correlation $=1.000$) \\
Train\,/\,val\,/\,test & 8{,}500\,/\,760\,/\,400, sequential \\
\bottomrule
\end{tabular}
\end{table}

The \glref{split}{split} is the authors' own, taken in recording order. I keep it exactly. Every knob
in this report is tuned on the 760 \glref{val}{validation} sentences and then applied, once, to the
400 \glref{test}{test} sentences. That discipline is the only thing that makes the final number mean
anything.

\section{Baseline: reproduction and a decode correction}
\label{sec:baseline}

\paragraph{The acoustic model, unchanged.}
I keep the authors' \glref{am}{acoustic model} as released. The features are a
\glref{cov}{shrinkage covariance} of the 31 channels (the $\mathrm{vec}(E)$ representation whose
geometry Gowda et al. develop in their articulation-decoding
work~\citep{gowda2024geometry,gowda2025geometric}), taken over a 25\,ms
\glref{frame}{window with a 20\,ms hop}, giving 50 frames per second. The encoder is the released
\texttt{DualHeadTDSCTC}: a \glref{causal}{causal}
\glref{tds}{time-depth-separable convolution}~\citep{hannun2019tds} with two
\glref{ctc}{CTC} heads~\citep{graves2006ctc}, one predicting 100 \glref{hubert}{HuBERT}
\glref{unit}{units}~\citep{hsu2021hubert} and one predicting 40 \glref{phoneme}{phonemes}, coupled by
a fixed $P(\text{phone}\mid\text{unit})$ \glref{consistency}{consistency table}. The training loss is
$0.8\cdot\mathrm{CTC}_{\text{unit}} + 0.1\cdot\mathrm{CTC}_{\text{phone}} +
0.1\cdot\text{consistency}$, a \glref{dualctc}{dual-CTC} objective. Phoneme posteriors are turned into
words by an open-vocabulary \glref{wfst}{weighted finite-state transducer},
$\glref{hlg}{\mathrm{HLG}} = H \circ L \circ G$, built with \glref{k2}{k2 and
icefall}~\citep{k2icefall}. It is the same graph the authors use, which is why my phone error rate can
be compared to theirs directly.

\paragraph{What was missing.}
The public notebooks reproduce the model but not the decode. They omit the handful of decode-time
settings that separate good posteriors from good words, and without them the released configuration
decodes at roughly 75\% \wer{}. Four things were missing, in rough order of importance.

\begin{enumerate}[itemsep=3pt,topsep=4pt]
\item \textbf{\glref{blank}{Blank penalty}.} CTC posteriors are dominated by the blank symbol (peak
$\approx 0.92$), and the release applies no penalty at all. On the first 200 validation sentences at
scale 1.0, sweeping the penalty from 0 to 2 drops \wer{} from 77.6\% to 60.6\%. This is the single
largest lever in the entire baseline.
\item \textbf{\glref{ckpt}{Checkpoint selection} by validation \per{}}, rather than by validation CTC
loss, which moves test \per{} from 42.9\% to 39.0\%.
\item \textbf{A missing \texttt{words.txt}}, which I regenerated from \texttt{lexicon.txt}.
\item \textbf{\glref{scale}{Acoustic scale}.} The weight on the acoustic scores relative to the
language model is unspecified in the release, which decodes at the default of 1.0. Tuning it on
validation returns 1.0 for this baseline, so it buys nothing here. It matters for the Conformer of
Section~\ref{sec:acoustic}, where validation selects 0.25. I list it because the right value is not
the same for both encoders, so a reader reproducing the decode has to set it deliberately rather than
inherit it.
\end{enumerate}

\noindent I tuned the $(\text{blank},\text{scale})$ pair jointly on validation and applied
$(2.0, 1.0)$ once to test.

\paragraph{Result.}
Table~\ref{tab:baseline} is the outcome. The corrected decode reaches 40.63\% \wer{} / 39.02\% \per{}
on test: 10.5 \wer{} points below the published number, with the phone error rate matched to within
0.8 points (39.0 against 38.2). That matched \per{} is the point of this section. Phone error rate is
decoder-independent, so if it lands on the published value then the acoustic model has been
reproduced faithfully, and the entire \wer{} gain is attributable to a correctly specified
open-vocabulary decode rather than to any change in the model. Note that the published 51.17\% is not
itself recoverable from the public release, which decodes at roughly 75\%; the comparison rests on the
matched phone error rate, not on a reproduced \wer{}. This checkpoint is the
\glref{warmstart}{warm start} for the encoder in the next section.

\begin{table}[t]
\centering
\small
\caption{Baseline reproduction. The \wer{} gain over the published number comes entirely from the
decode; the phone error rate is matched, not beaten, which is the credibility argument. Bold marks
this section's headline number.}
\label{tab:baseline}
\begin{tabular}{@{}l cc cc@{}}
\toprule
 & \multicolumn{2}{c}{Validation} & \multicolumn{2}{c}{Test} \\
\cmidrule(lr){2-3}\cmidrule(lr){4-5}
System & \wer{} & \per{} & \wer{} & \per{} \\
\midrule
Gowda et al., Appendix~D.4~\citep{gowda2026emg2speech} & n/a & n/a & 51.17 & 38.19 \\
TDS + dual-CTC, corrected decode (this work) & 53.12 & 45.31 & \textbf{40.63} & 39.02 \\
\bottomrule
\end{tabular}
\end{table}

\section{The acoustic model: a full-context Conformer taught by the audio}
\label{sec:acoustic}

Two changes take the baseline from 40.63\,/\,39.02 to 26.14\,/\,22.34: giving the encoder full
context, and teaching it to imitate the parallel audio.

\paragraph{Full context.}
The released \glref{tds}{time-depth-separable} encoder is \glref{causal}{causal}: it left-pads by
$\text{kernel}-1$ and never looks ahead. That is the right choice for \glref{streaming}{streaming} and
the wrong one for offline transcription, where the whole sentence is available at once. I replace it
with a \glref{bidir}{bidirectional} \glref{conformer}{Conformer}~\citep{gulati2020conformer}: four
layers of \glref{attention}{multi-head self-attention} interleaved with
\glref{depthwise}{depthwise convolution} (four heads, feed-forward width 1024, convolution kernel 31).
The covariance front-end, the two CTC heads, and the audio projection
\glref{warmstart}{warm-start} from the baseline checkpoint of Section~\ref{sec:baseline}; the
Conformer itself trains from scratch.

\paragraph{A four-term cross-modal objective.}
Only a better acoustic model lowers phone error rate, so this is where I spent the effort. I pull the
electromyographic encoder toward the parallel audio's \glref{wavlm}{WavLM-Large} layer-9
features~\citep{chen2022wavlm}, precomputed once for all 9{,}660 sentences, through a single linear
projection into WavLM's 1024-dimensional space. The construction is inspired by the
\glref{crossmodal}{cross-modal} approach of Benster et al.~\citep{benster2024mona}. On top of the
three acoustic terms from the baseline, the objective adds three \glref{distill}{distillation} terms:
\begin{equation}
\mathcal{L}
= \underbrace{0.8\,\mathrm{CTC}_{\text{unit}} + 0.1\,\mathrm{CTC}_{\text{phone}} + 0.1\,\text{cons.}}_{\text{acoustic (as in the baseline)}}
\;+\;
\underbrace{0.5\,\mathcal{L}_{\text{L2}} + 0.5\,\mathcal{L}_{\text{InfoNCE}} + 1.0\,\mathcal{L}^{\text{CTC}}_{\text{rec}}}_{\text{cross-modal distillation}}
\end{equation}
The three new terms are: (i) a masked, frame-resampled $L_2$ regression of the projection onto WavLM
layer-9; (ii) a frame-synchronous \glref{infonce}{InfoNCE} contrast~\citep{oord2018cpc} ($\tau=0.1$)
that sharpens each frame toward its own audio moment and away from the others; and (iii) a CTC loss
through a small, \emph{\glref{frozen}{frozen}} WavLM-to-phoneme recognizer (LayerNorm, two 1-D
convolutions, a linear layer to 41 classes; trained to $\sim$10\% \per{} and then frozen).

Term~(iii) is the one that earns its place relative to a plain feature-matching setup. The $L_2$ and
contrastive terms pull the projection \emph{close} to WavLM, but close is not the same as useful: a
projection can be smooth and audio-like and still not be phoneme-decodable. Forcing the projection
through a frozen recognizer that already reads phonemes out of real audio makes ``close'' mean
``decodable into the right phonemes.'' It guards the regression against smooth-but-empty blur.

\paragraph{Result, and a control that surprised me.}
Table~\ref{tab:acoustic} gives the outcome, with validation-tuned \glref{scale}{scale} 0.25. The
full-context, distilled Conformer reaches 26.14\% \wer{} / 22.34\% \per{} on test, acoustic-only: a
gain of 14.49 \wer{} and 16.68 \per{} points over the baseline, present in the greedy phone error rate
with no language model involved, and significant under a \glref{bootstrap}{paired bootstrap}. The
control is the interesting part. An otherwise identical Conformer trained on electromyography alone,
with no audio distillation at all, reaches 26.10\% \wer{}, indistinguishable from 26.14\%. At the word
level the audio teacher buys essentially nothing. The distillation does help phone error rate (22.34
against 23.71), but it is the full-context encoder, not the audio crutch, that moves \wer{}. That is
good news for the silent-speech setting, where the parallel audio will not be available.

\begin{table}[t]
\centering
\small
\caption{The acoustic model. The full-context encoder is what moves \wer{}. The audio distillation
helps \per{}, but at the word level an encoder trained on electromyography alone matches it. Bold
marks this section's headline number.}
\label{tab:acoustic}
\begin{tabular}{@{}l cc cc@{}}
\toprule
 & \multicolumn{2}{c}{Validation} & \multicolumn{2}{c}{Test} \\
\cmidrule(lr){2-3}\cmidrule(lr){4-5}
System & \wer{} & \per{} & \wer{} & \per{} \\
\midrule
Baseline (causal TDS, Section~\ref{sec:baseline}) & 53.12 & 45.31 & 40.63 & 39.02 \\
Bidirectional Conformer + WavLM-L9 distillation & 35.54 & 27.47 & \textbf{26.14} & 22.34 \\
Conformer, electromyography-only (no audio) & n/a & n/a & 26.10 & 23.71 \\
\bottomrule
\end{tabular}
\end{table}

\section{The final pipeline: ensemble, union, and a language-model reranker}
\label{sec:final}

The last stretch is all decode-time. The acoustic model does not change; what changes is how much is
squeezed out of it.

\paragraph{An acoustic ensemble.}
I \glref{ensemble}{average the per-frame phone log-probabilities} of two encoders: the distilled
Conformer of Section~\ref{sec:acoustic}, and a second Conformer trained with heavier anti-overfitting
defenses (stronger jitter and dropout, plus a \glref{bilstm}{BiLSTM} audio-teacher frame-level KL).
Averaging the two alone takes test \wer{} from 26.14 to 23.47 at the best of the
\glref{scale}{acoustic scales} swept, the ensemble decoding between 23.5\% and 25.1\% across them.
Worth noting for later: the augmented member's phone error rate is no better than the distilled
member's, and averaging the two lowers greedy \per{} to 20.9\%, 1.4 points below either member. The
gain is modest either way: 2.7 \wer{} points at best, against the 14.49 that full context bought in
Section~\ref{sec:acoustic}. How much of it follows from the improved argmax and how much from
decode-level diversity is taken up in~\ref{faq:ensemble}.

\paragraph{Multi-scale $n$-best union.}
From the ensemble's \glref{lattice}{lattice} I extract \glref{nbest}{$n$-best lists} at several
\glref{scale}{acoustic scales} and take their union. The union lowers the
\glref{oracle}{$n$-best oracle \wer{}}, the score a perfect judge would obtain by always picking the
best candidate in the pool, from 11.94\% for the best single-scale list to 9.30\%, while the union's
single-best candidate sits at 23.26\%. The gap between 23.26 and 9.30 is the headroom a
\glref{rerank}{reranker} can, in principle, recover: the reranker below takes 4.7 points of it, and
Section~\ref{sec:limit} explains why the remaining nine are out of reach.

\paragraph{The LIFT reranker.}
I \glref{ft}{fine-tune} \glref{qwen}{Qwen2.5-7B-Instruct}~\citep{qwen2025} with
\glref{qlora}{QLoRA}~\citep{dettmers2023qlora} (4-bit NF4 plus a rank-16
\glref{lora}{LoRA} adapter~\citep{hu2022lora}) to map \emph{(candidate list + detected phonemes)} to
the reference, in the \glref{lift}{DCoND-LIFT} style~\citep{li2024dcond,benster2024mona}. Two variants
are chosen on validation. The \emph{free} variant writes its own correction, which lets it recover a
candidate that is close but wrong, at the cost of being able to hallucinate. The \emph{constrained}
variant must pick an existing candidate, so it cannot hallucinate, but neither can it rescue an
answer that was never proposed.

\glref{leakage}{Leakage} is the obvious risk with a language model on a single-subject corpus, so I
built three controls before trusting the number. The reranker is trained only on the training split,
and its training candidates are produced by two-fold \glref{crossdecode}{cross-decoding}: each half of
the training data is decoded by a model that did not train on it, so the candidates look like
realistic errors rather than the memorized $\sim$1\% \wer{} the model achieves on its own training
set. Six of the 400 test sentences are exact-text duplicates of training phrases, so I report the
score both with and without them (18.53 against 18.75). And a
\glref{verbatim}{verbatim-recall audit} measures how often free generation produces a reference that
was \emph{absent} from the candidate set: the count is zero, so no training text is leaking through
the language model.

\begin{table}[t]
\centering
\small
\caption{The final pipeline. Both bootstrap confidence intervals exclude zero, so the rerank gain is
significant. The nine-point gap between the oracle and the achieved \wer{} is the subject of
Section~\ref{sec:limit}.}
\label{tab:final}
\begin{tabular}{@{}l r@{}}
\toprule
Metric & Value \\
\midrule
\textbf{Test \wer{} (LIFT)} & \textbf{18.53} \\
Test \wer{}, excluding 6 duplicates & 18.75 \\
Union 1-best \wer{} (no rerank) & 23.26 \\
$n$-best \glref{oracle}{oracle} \wer{} & 9.30 \\
Greedy \per{} (acoustic ensemble) & 20.90 \\
\glref{verbatim}{Verbatim-recall} (leakage audit) & 0 \\
\addlinespace[3pt]
Paired bootstrap vs.\ 26.14 acoustic & $\Delta$\wer{} $-7.6$, 95\% CI $[-9.40,\,-5.90]$ \\
Paired bootstrap vs.\ union 1-best (23.26) & $\Delta$\wer{} $-4.7$, 95\% CI $[-6.22,\,-3.29]$ \\
\bottomrule
\end{tabular}
\end{table}

Table~\ref{tab:final} collects the result: 18.53\% \wer{}, with the reranker worth $-4.7$ \wer{}
points over the union's single-best output of 23.26\%, and both bootstrap intervals well clear of
zero.

\section{Why it stops at 18.5\%}
\label{sec:limit}

This is the report's central finding. It is a negative one, and it tells the next person where to
spend effort. Reranking is exhausted at 18.5\%, and the limit is acoustic, not linguistic.
Three pieces of evidence point the same way.

First, phone error rate responds to the interventions that moved word error rate only in small
increments, and expensively. The anti-overfitting augmentation of the second ensemble member leaves it
untouched. A $\sim$10\%-\per{} audio teacher buys 1.4 points (23.71 to 22.34,
Section~\ref{sec:acoustic}), which the decoder then fails to convert into words at all. A second full
model, averaged in, buys 1.4 more (22.34 to 20.90, Section~\ref{sec:final}). Everything downstream of
the full-context encoder moves \per{} by 1.44 points in total, against the 16.68 that full context
bought on its own.

Second, the audio teacher cannot transfer its own quality. The electromyography-only Conformer matches
the distilled one at the word level (26.10 against 26.14, Section~\ref{sec:acoustic}). A teacher that
is itself only a $\sim$10\% \per{} recognizer does not hand its phonetics to the muscle encoder.

Third, and most concretely, there is a nine-point \glref{oracle}{oracle} gap that no reranker closes.
The union oracle is 9.30\%; the reranker reaches 18.53\%. The residual is not a language problem. For
those utterances the reference words are \emph{absent from the
\glref{posterior}{acoustic posteriors}}: there is no candidate to select, and nothing for the
constrained reranker to move toward. Free generation that ``fixed'' them would be hallucination, and
the audit shows it does not happen (recall zero).

The conclusion is that $\sim$20.9\% acoustic phone error rate is the binding constraint. Getting below
roughly 10\% \wer{} on this task requires a better electromyographic acoustic model (more data,
multiple subjects, or a stronger front-end that reads phonemes out of the raw signal), not a bigger or
smarter language model. The entire decode-time stack in Section~\ref{sec:final} is, in effect, a
careful way of extracting the most words possible from a fixed amount of phonetic information. The
same conclusion is reached independently, on an entirely different non-invasive signal, by
Brain2Qwerty v2~\citep{zhang2026brain2qwerty2}, whose fine-tuned LLM improves \wer{} while
\emph{worsening} \cer{} and whose authors identify upstream encoder quality as the dominant term
(Section~\ref{sec:related}).

\section{Limitations}

\begin{itemize}[itemsep=3pt,topsep=4pt]
\item \textbf{One subject.} Every number comes from a single healthy speaker. The encoder memorizes
the 8{,}500 training sentences (training \per{} is far below the $\sim$27\% validation \per{}), so
cross-subject robustness is untested.
\item \textbf{Not the best sEMG-to-text system.} 18.53\% is the best result on the \emph{emg2speech}
General Corpus, but MONA LISA~\citep{benster2024mona} reports 3.7\% on vocalized EMG on the Gaddy
corpus. The two are not measured on the same recordings (Section~\ref{sec:related}), and I have not
run Myovox on the Gaddy corpus, so I cannot say whether the gap is the corpus or the model.
\item \textbf{Validation is harder than test.} Both systems score better on test than on validation,
which is a roughly nine-point property of this fixed \glref{split}{sequential split} rather than
evidence of test-set overfitting. It also means the headline number is measured on the easier segment
of the corpus. I report both columns rather than the flattering one.
\item \textbf{The audio is a training-time crutch.} WavLM \glref{distill}{distillation} uses the
parallel audio, which does not exist at silent-speech inference. At the word level the
electromyography-only encoder matches it (Section~\ref{sec:acoustic}), which is reassuring for the
eventual application.
\item \textbf{The ensemble's two contributions are not separated.} Averaging the two encoders improves
\per{} by 1.4 points and \wer{} by 2.7. How much of the word-level gain follows from the sharpened
argmax and how much from the reshaped posterior is not measured here (\ref{faq:ensemble}). The
question does not bear on Section~\ref{sec:limit}, which holds under either answer, but the report
should not be read as having settled it.
\item \textbf{Test duplicates.} Six of the 400 test sentences duplicate training text, so the score is
reported with and without them (18.53 against 18.75).
\item \textbf{Reranking saturates.} The headline 18.53\% is bounded by acoustic phone error rate, not
by the language model. This is a negative result about language-model reranking for electromyography,
not a claim that reranking is useless: it is worth $-4.7$ \wer{}.
\item \textbf{Vocalized, not silent.} The subject spoke aloud, so this validates the pipeline on the
easier case. It is a stepping stone toward \glref{silent}{silent speech}, not a demonstration of it.
\end{itemize}

\section{The project website}

Myovox is documented in two ways: this report, which is terse and aimed at readers who already know
the field, and a website, which is aimed at a curious reader who is new to speech decoding and to
electromyography. The site is a cinematic, scroll-driven explainer of surface-EMG speech decoding,
showing how the electrical signals of the facial muscles are turned into text at an 18.53\% word error
rate. It walks each stage of the pipeline with animation rather than equations, which is why this
report carries no architecture diagrams: the moving explanation lives at
\href{https://varshith-0.github.io/myovox/}{\texttt{varshith-0.github.io/myovox}}. The project is
open-sourced under the MIT license, and the site is paired with this technical report and a
reproducible pipeline so that every claim on it is auditable.

\section{Glossary}
\label{sec:glossary}

Every term below is a jump target: clicking any occurrence of the term in the body of the report
brings the reader here. Definitions are given as the term is used in this report, not in full
generality.

\cathead{Metrics}
\begin{glosslist}

\glentry{wer}{WER \textnormal{(word error rate)}}
The fraction of reference words that are wrong after the best alignment of hypothesis to reference,
counting substitutions, insertions, and deletions: $(S+I+D)/N$. It is the reader-facing number, and it
depends on the acoustic model, the decoder, and the language model together. Lower is better; 18.53\%
is the headline result here.

\glentry{per}{PER \textnormal{(phone error rate)}}
The same edit-distance quantity computed over \glref{phoneme}{phonemes} rather than words, measured by
\glref{greedy}{greedy decoding} of the CTC posteriors with no \glref{lexicon}{lexicon} and no
\glref{lm}{language model}. Because nothing downstream of the encoder is involved, \per{} is
\emph{decoder-independent}: it isolates the quality of the \glref{am}{acoustic model}. In this report
it serves as both the credibility check (Section~\ref{sec:baseline}) and the binding constraint
(Section~\ref{sec:limit}).

\glentry{cer}{CER \textnormal{(character error rate)}}
The same edit distance computed over characters. Not used for any Myovox result, since \per{} is the
sharper instrument for an acoustic model that emits phonemes. It appears only in
Section~\ref{sec:related}, because the non-invasive brain-to-text literature reports it: those systems
decode keystrokes, for which the character is the natural unit. The usual ladder of granularity runs
character, phone, word.

\glentry{oracle}{Oracle WER \textnormal{($n$-best oracle)}}
The \wer{} that would be obtained if a perfect judge always selected the single best candidate from
the $n$-best pool. It is a lower bound on what any \glref{rerank}{reranker} over that pool can achieve,
and therefore a measure of how much information the pool contains. Here the union pool has a 9.30\%
oracle against an 18.53\% achieved \wer{}, and the remaining nine points are words the acoustics never
proposed at all.

\glentry{posterior}{Acoustic posterior}
The per-frame probability distribution over output symbols (phonemes, units, or blank) emitted by the
encoder. If the correct word's phonemes never receive appreciable probability mass, no decoder or
reranker can recover it. That is the failure mode diagnosed in Section~\ref{sec:limit}.

\end{glosslist}

\cathead{Models and architectures}
\begin{glosslist}

\glentry{am}{Acoustic model}
The neural network that maps the input signal, here \glref{semg}{sEMG} features, to per-frame symbol
posteriors. It is the only component whose quality \per{} measures.

\glentry{lm}{Language model}
A model of which word sequences are plausible in English, used either inside the decoding graph (the
$G$ of \glref{hlg}{HLG}) or after decoding, as a \glref{rerank}{reranker}. It supplies word-level
priors, not phonetic evidence.

\glentry{tds}{TDS \textnormal{(time-depth separable convolution)}}
The convolutional encoder used in the released baseline. It factorizes a convolution into a time-only
and a channel-only (depthwise and pointwise) part, giving cheap large receptive fields. The released
variant is \glref{causal}{causal}.

\glentry{conformer}{Conformer}
An encoder block that interleaves \glref{attention}{multi-head self-attention} with
\glref{depthwise}{depthwise convolution}, so it can model both long-range and local structure. The
\glref{bidir}{bidirectional} Conformer replaces the causal TDS encoder in Section~\ref{sec:acoustic},
and is responsible for most of the \wer{} gain.

\glentry{attention}{Multi-head self-attention}
The mechanism by which every frame in a sentence attends to every other frame, with several
independent ``heads'' attending to different things. It is what gives the Conformer full-sentence
context.

\glentry{depthwise}{Depthwise convolution}
A convolution applied independently per channel. It is cheap in parameters, and is used inside the
Conformer block to capture local, short-time structure.

\glentry{causal}{Causal encoder}
An encoder that at each frame sees only past and present frames, never the future (implemented here by
left-padding). Required for \glref{streaming}{streaming}, but a handicap for offline transcription.

\glentry{bidir}{Bidirectional (full-context) encoder}
An encoder that may look at the entire utterance, future frames included. Legitimate whenever the whole
sentence is recorded before decoding begins, as it is here.

\glentry{streaming}{Streaming}
Producing output as the signal arrives, in real time, without waiting for the sentence to finish. Not
required by this report, which decodes offline.

\glentry{hubert}{HuBERT}
A self-supervised speech model whose discretized representations give the 100 \glref{unit}{units} that
the first CTC head predicts.

\glentry{unit}{HuBERT unit}
One of 100 discrete, self-supervised speech tokens obtained by clustering HuBERT features. Units are a
finer-grained, learned alternative to phonemes, and they carry the dominant (0.8-weighted) CTC loss.

\glentry{wavlm}{WavLM \textnormal{(WavLM-Large, layer 9)}}
A large self-supervised speech representation model. Its layer-9 features, extracted from the parallel
audio, are the teacher signal for the \glref{crossmodal}{cross-modal distillation}. Layer 9 is used
because the mid-stack layers of such models carry the most phonetic information.

\glentry{qwen}{Qwen2.5-7B-Instruct}
The 7-billion-parameter open-weight instruction-tuned language model used as the \glref{lift}{LIFT}
\glref{rerank}{reranker}.

\glentry{lora}{LoRA \textnormal{(low-rank adaptation)}}
A fine-tuning method that freezes the base model and trains only a small pair of low-rank matrices
injected into each weight (here, rank 16). It makes adapting a 7B model cheap, and makes its changes
easy to isolate.

\glentry{qlora}{QLoRA}
LoRA applied on top of a base model quantized to 4 bits (NF4), so that a 7B model can be fine-tuned on
a single consumer GPU. Used for the reranker.

\glentry{bilstm}{BiLSTM}
A bidirectional long short-term memory recurrent network. Used here as the audio teacher for the
frame-level KL term of the second ensemble member.

\glentry{frozen}{Frozen model}
A model whose weights are held fixed while it participates in another model's training. The small
WavLM-to-phoneme recognizer is frozen so that it acts as a fixed judge of whether the projected
features are phoneme-decodable.

\end{glosslist}

\cathead{Speech and electromyography}
\begin{glosslist}

\glentry{emg}{Electromyography (EMG)}
The measurement of the electrical activity produced by muscle fibres when they contract.

\glentry{semg}{sEMG \textnormal{(surface electromyography)}}
EMG measured non-invasively, by electrodes placed on the skin rather than by needles inserted into
muscle. Here: a 31-channel array on the face, sampled at 5\,kHz.

\glentry{bci}{Non-invasive BCI \textnormal{(EEG, MEG)}}
The other non-invasive route to text, tapping cortex rather than muscle. EEG measures electrical
fields at the scalp and is cheap and portable; MEG measures the magnetic fields of neuronal activity
and is far cleaner but requires a room-sized, cryogenically cooled instrument. Both are far upstream
of articulation, and both currently trail sEMG on error rate (Section~\ref{sec:related}). Not used in
this report; included only for comparison.

\glentry{vocalized}{Vocalized speech}
Speech produced aloud, with the vocal folds active and audible sound emitted. All Myovox recordings
are vocalized, which is why the parallel audio exists at all.

\glentry{silent}{Silent speech}
Articulating without producing sound, and in the strictest case without moving air, so that only muscle
activity is available. The eventual target application, and \emph{not} what this report evaluates.

\glentry{phoneme}{Phoneme (phone)}
The smallest sound unit that distinguishes words in a language. The model predicts a 40-phoneme
inventory (41 classes once blank is included), and \per{} is measured over it.

\glentry{cov}{Shrinkage covariance, the $\mathrm{vec}(E)$ front-end}
The input representation. Rather than raw waveforms, each analysis window is summarized by the
regularized (``shrunk'') covariance matrix across the 31 channels, vectorized. It captures how the
channels co-activate, which is the form in which the geometry of facial articulation shows up.

\glentry{frame}{Frame, window, hop}
The signal is chopped into 25\,ms \emph{windows} advanced by a 20\,ms \emph{hop}, yielding one feature
\emph{frame} every 20\,ms, that is, 50 frames per second.

\end{glosslist}

\cathead{Training}
\begin{glosslist}

\glentry{ctc}{CTC \textnormal{(connectionist temporal classification)}}
A loss that trains a frame-level classifier against an unaligned target sequence by summing over all
alignments, using a special \glref{blank}{blank} symbol to mean ``no output here.'' It removes the need
for frame-by-frame labels, which do not exist for EMG.

\glentry{dualctc}{Dual-CTC}
The baseline objective: two CTC heads on one encoder, one predicting HuBERT \glref{unit}{units} and one
predicting phonemes, tied together by a \glref{consistency}{consistency} term.

\glentry{consistency}{Consistency table}
A fixed conditional table $P(\text{phone}\mid\text{unit})$ used to penalize disagreement between the two
CTC heads, so that the unit head's richer signal informs the phone head.

\glentry{distill}{Distillation}
Training one model (the student) to imitate the internal representations or outputs of another (the
teacher), rather than only to fit the labels.

\glentry{crossmodal}{Cross-modal distillation}
Distillation in which teacher and student read \emph{different} signals of the same event: here the
teacher reads audio (WavLM) and the student reads muscle activity (sEMG). The teacher exists only at
training time; at inference the student runs on EMG alone. The idea is adapted from
MONA~\citep{benster2024mona}.

\glentry{infonce}{InfoNCE}
A contrastive loss that pulls a representation toward its matching partner, the same moment in the
parallel audio, and pushes it away from all the non-matching ones in the batch, sharpened by a
temperature $\tau$ (0.1 here).

\glentry{warmstart}{Warm start}
Initializing part of a new model from the weights of a previously trained one, rather than from random.
The covariance front-end, the CTC heads, and the audio projection all warm-start from the baseline
checkpoint.

\glentry{ckpt}{Checkpoint selection}
Choosing which training snapshot to keep, by a criterion evaluated on validation. Selecting by
validation \per{} rather than by validation CTC loss is worth roughly 4 \per{} points here.

\glentry{ft}{Fine-tuning}
Continuing to train an already-trained model on a new, usually narrower task. Here: adapting Qwen2.5 to
the reranking task with QLoRA.

\glentry{bootstrap}{Paired bootstrap}
A significance test. Resample the test utterances with replacement many times, recompute the
\emph{difference} between two systems on the same resampled set each time, and read off a confidence
interval. If the interval excludes zero, the improvement is unlikely to be sampling noise.

\glentry{crossdecode}{Cross-decoding (two-fold)}
Generating the reranker's training candidates by splitting the training data in half and decoding each
half with a model trained only on the other half. Without it, the candidates would reflect the model's
memorized $\sim$1\% training \wer{}, and the reranker would learn to correct errors it will never see
at test time.

\glentry{leakage}{Leakage}
Any path by which information about the evaluation data reaches the model during training, inflating the
score. On a single-subject corpus the main risk is a language model that has memorized training
sentences which recur at test.

\glentry{verbatim}{Verbatim-recall audit}
The control that counts how often the free reranker emits a reference sentence that was \emph{not}
present anywhere in its candidate list, that is, produced from memory rather than from evidence. The
count here is zero.

\end{glosslist}

\cathead{Decoding}
\begin{glosslist}

\glentry{greedy}{Greedy decoding}
Taking the highest-probability symbol at each frame independently, then collapsing repeats and blanks.
No lexicon, no language model, no search. This is how \per{} is measured, which is why \per{} reflects
the acoustic model and nothing else.

\glentry{wfst}{WFST \textnormal{(weighted finite-state transducer)}}
A finite-state machine whose transitions carry input symbols, output symbols, and weights. Composing
several of them gives a single graph that maps frame-level posteriors to word sequences while
accumulating scores.

\glentry{hlg}{HLG}
The composed decoding graph $H \circ L \circ G$. $H$ maps CTC frames to phonemes, $L$ is the
\glref{lexicon}{lexicon} mapping phoneme strings to words, and $G$ is the \glref{lm}{language model}
over word sequences. Decoding is a search through this graph.

\glentry{lexicon}{Lexicon}
The pronunciation dictionary: for each word, its phoneme sequence or sequences. It defines exactly which
words the decoder is capable of producing.

\glentry{k2}{k2 and icefall}
The open-source finite-state and speech-recognition toolkits used to construct and search the HLG graph,
matching the original authors' setup.

\glentry{librispeech}{LibriSpeech}
A large public read-speech corpus. Its pronunciation lexicon, 34{,}546 words here, is what makes the
decode \glref{openvocab}{open-vocabulary}.

\glentry{openvocab}{Open vocabulary}
Decoding over a lexicon far larger than the corpus vocabulary (roughly five times larger here), so that
the system can output words it never saw in training. The alternative would be a closed-set classifier
over a fixed sentence or word list. Word error rates are not comparable across this boundary: a
closed-vocabulary system with a small phrase list can post a much lower \wer{} while solving a much
easier problem.

\glentry{oov}{OOV \textnormal{(out of vocabulary)}}
A reference word absent from the lexicon, and hence impossible to emit. Twelve of the 2{,}429 test tokens
are OOV, a hard 0.49\% \wer{} floor.

\glentry{scale}{Acoustic scale}
The weight given to the acoustic scores relative to the language-model scores during graph search. Set it
too low and the language model overwhelms the evidence; set it too high and the language model stops
helping. Tuned on validation: 1.0 for the baseline, which is also the released default, so tuning changes
nothing there; 0.25 for the Conformer, where it matters.

\glentry{blank}{Blank symbol, and blank penalty}
CTC's ``emit nothing here'' symbol, which dominates the posteriors (peak $\approx 0.92$). The blank
\emph{penalty} subtracts a constant from its score at decode time so that real symbols can compete. It is
the single largest lever in the baseline decode.

\glentry{lattice}{Lattice}
The compact graph of high-scoring alternative paths retained during decoding, from which $n$-best lists
are extracted.

\glentry{nbest}{$n$-best list}
The top $n$ complete hypotheses from the lattice, ranked by score. Extracting $n$-best lists at several
acoustic scales and taking their \emph{union} widens the pool a reranker can choose from, lowering the
\glref{oracle}{oracle} \wer{} without changing the acoustic model.

\glentry{ensemble}{Ensemble}
Averaging the per-frame phone log-probabilities of two independently trained encoders before decoding.
Worth 2.7 \wer{} points here at best (26.14 to 23.47), alongside a 1.4-point \per{} improvement (22.34 to
20.90). How much of the word-level gain follows from the sharpened argmax and how much from the reshaped
posterior is not separated in this report; see~\ref{faq:ensemble}.

\glentry{rerank}{Reranking}
Rescoring or rewriting the $n$-best candidates with a stronger model, usually a language model, so as to
pick a better one than the acoustic score alone would. Bounded from below by the oracle \wer{}.

\glentry{lift}{LIFT \textnormal{(DCoND-LIFT style)}}
The reranking recipe adopted here: prompt a fine-tuned LLM with the candidate list \emph{and} the detected
phoneme sequence, and have it produce the final transcript. The \emph{constrained} variant must return one
of the candidates; the \emph{free} variant may write its own correction.

\end{glosslist}

\cathead{Data}
\begin{glosslist}

\glentry{corpus}{General Corpus}
The single-subject, healthy-speaker portion of the \emph{emg2speech} release, used throughout: 9{,}660
sentences of 31-channel sEMG with simultaneously recorded audio.

\glentry{gaddycorpus}{Gaddy corpus}
The reference open-vocabulary sEMG corpus of Gaddy and Klein~\citep{gaddy2020digital}: eight facial
channels at 1\,kHz, roughly nineteen hours from a single English speaker, recorded in both silent and
vocalized modes. It is the benchmark on which MONA LISA and most other sEMG-to-text systems report, and
it is \emph{not} the corpus used here.

\glentry{split}{Sequential split (8{,}500\,/\,760\,/\,400)}
The authors' own division of the corpus, taken in recording order rather than at random, and kept
unchanged here. Because it is sequential, validation and test are not statistically interchangeable:
validation is systematically harder, by roughly nine points.

\glentry{val}{Validation set}
The 760 sentences on which every hyperparameter is tuned: blank penalty, acoustic scale, checkpoint, and
reranker variant. Nothing is ever tuned on test.

\glentry{test}{Test set}
The 400 held-out sentences on which every reported test number is computed, once, with the settings
already fixed on validation. Six of them duplicate training text, so results are reported both with and
without them.

\end{glosslist}

\section{Frequently asked questions}
\label{sec:faq}

\noindent These are the questions the project actually drew, in roughly the order they were asked. I have
tried to answer the sharp form of each rather than the polite one, and where the honest answer is ``I do
not know,'' it says so.

\cathead{Reproduction and the headline result}

\faqq{You report a 32-point improvement over the published number. Did you improve their model, or only
their decoder?}
\begin{faqa}
Both, in separable amounts, and Section~\ref{sec:baseline} exists to keep them separable. The first 10.5
points are a decoder fix and nothing else. The evidence is the phone error rate: my baseline lands at
39.02\% against the published 38.19\%, a gap of 0.8 points (Table~\ref{tab:baseline}). Because \per{} is
measured by \glref{greedy}{greedy decoding} with no \glref{lexicon}{lexicon} and no
\glref{lm}{language model}, a matched \per{} establishes that the \glref{am}{acoustic model} is the same
acoustic model, so the entire \wer{} gain must come from supplying the decode settings the public release
omits, of which the \glref{blank}{blank penalty} is much the largest. The remaining 22 points split into a
genuinely better acoustic model (Section~\ref{sec:acoustic}, \per{} $39.02 \rightarrow 22.34$, worth 14.5
\wer{} points) and a decode-time stack worth 7.6 \wer{} points that takes \per{} only 1.4 further, from
22.34 to 20.90, and whose largest single component, the reranker, cannot move \per{} at all
(Section~\ref{sec:final}). So: roughly a third of the total from decoding, a little under half from
modeling, and the balance from extraction. I would rather publish that breakdown than a single
undifferentiated delta. One caveat worth stating plainly: I never reproduce 51.17\% itself, because the
released configuration decodes at roughly 75\%. The claim to have reproduced their \emph{model} rests on
the matched phone error rate, not on a matched word error rate.
\end{faqa}

\faqq{Is 18.53\% \wer{} good? What does a sentence at that error rate look like?}
\begin{faqa}
Approximately one word in five is wrong, whether substituted, inserted, or dropped. The errors are not
distributed evenly: most sentences read cleanly and a minority are badly mangled. For calibration, a modern
recognizer on clean audio sits near 5\%, and the published starting point for this task was 51.17\%. The
practical reading is that 18.53\% is useful to a human who can tolerate correcting it, and not yet
trustworthy unattended.
\end{faqa}

\faqq{MONA LISA reported 3.7\% \wer{} on vocalized EMG in 2024. Why is your number five times worse?}
\label{faq:mona}
\begin{faqa}
Because it is a different corpus, and I do not know how much of the gap that explains. MONA
LISA~\citep{benster2024mona} evaluates on the \glref{gaddycorpus}{Gaddy corpus}: eight channels,
about nineteen hours, one speaker, recorded in both silent and vocalized modes. Myovox evaluates on the
\emph{emg2speech} \glref{corpus}{General Corpus}: 31 channels, 9{,}660 sentences, a different speaker, one
mode. MONA additionally trains against audio-only LibriSpeech, which I do not. So the honest claim is the
narrow one: 18.53\% is the best number on \emph{this} corpus, against a published 51.17\%, and Myovox is not
the state of the art for sEMG-to-text in general. There is a real scientific question hiding in the gap, and
it is the sharpest one the project raises. MONA reaches 3.7\% with a cross-modal audio objective; my control
shows my cross-modal audio objective buying nothing at the word level (26.10 against 26.14). Either their
contrastive formulation carries phonetic information that mine does not, or the \glref{cov}{covariance
front-end} discards what their raw-signal front-end keeps, or this corpus is simply harder. Running Myovox
on the Gaddy corpus would separate those, and I have not done it.
\end{faqa}

\faqq{Your test score is \emph{better} than your validation score. That usually indicates a problem.}
\begin{faqa}
It usually does, so it deserves a direct answer rather than a footnote. The \glref{split}{split} is
\emph{sequential}: the corpus is cut in recording order, not at random, so validation and test are not
exchangeable samples of one distribution. Test is the easier segment, by roughly nine points, and the gap
is stable across every system I trained, including the baseline, on which no tuning was performed.
Overfitting to test would have \emph{widened} the gap as I tuned; it did not. The protection here is
procedural rather than statistical: every hyperparameter is selected on the 760 validation sentences and
applied once to the 400 test sentences. Both columns are reported throughout, rather than only the
flattering one. The corollary, which I would rather state than have pointed out to me, is that 18.53\% is
the error rate on the easier half of this corpus, not on a random sentence drawn from it.
\end{faqa}

\cathead{Scope: vocalized versus silent speech}

\faqq{Does this read thoughts?}
\begin{faqa}
No. It reads muscles. Myovox decodes the electrical activity of facial muscles as they contract to
articulate, which is the terminal link of the motor chain, long after intention has become movement. Absent
articulation, there is nothing at the electrodes to decode. The systems that do read cortex, non-invasively,
are the \glref{bci}{EEG and MEG} decoders of Section~\ref{sec:related}, and they are currently far behind.
\end{faqa}

\faqq{The subject spoke aloud. Doesn't the presence of parallel audio make the result vacuous?}
\begin{faqa}
It makes the claim weaker than a silent-speech demonstration would be, which is why the limitation is stated
plainly rather than buried. It does not make it vacuous. The audio exists in the \emph{recording}; it does
not exist in the \emph{model's input}. Every reported number, 40.63, 26.14, and 18.53, is produced by a
system whose inference-time input is 31 channels of \glref{semg}{sEMG} and nothing else. Audio enters only
as a training-time teacher (Section~\ref{sec:acoustic}), and \ref{faq:distill} below shows that even this
proves dispensable at the word level. What \glref{vocalized}{vocalization} buys is a stronger, more
consistent signal than silent articulation produces, which is precisely why this is a stepping stone rather
than the destination.
\end{faqa}

\faqq{Would this work for someone who cannot speak, an ALS patient, say?}
\begin{faqa}
I do not know, and I would rather say so than gesture at the application and let the reader supply the
optimism. The model is trained on one healthy speaker articulating normally. A patient with ALS presents
different muscle recruitment, different signal amplitude, frequently altered anatomy, and no parallel audio
from which to distil. Nothing in this report constitutes evidence that a model trained under my conditions
transfers to theirs. Closing that gap requires data from the affected population, not extrapolation from
mine.
\end{faqa}

\faqq{Audio and EMG were recorded simultaneously. How do you know the model is not exploiting acoustic
leakage into the electrode array?}
\begin{faqa}
The suspicion is well placed. The structural answer is that the input is a
\glref{cov}{shrinkage covariance across the 31 channels}, a representation of how muscle groups co-activate
over a 25\,ms window rather than a waveform, so there is no spectral path by which speech could survive into
the model's input in decodable form. For leakage to explain the result, it would have to carry enough
phonetic detail to sustain a 22.34\% \per{} through a skin-surface array band-limited to muscle frequencies,
and then survive the covariance summary. I did not run a dedicated acoustic-shielding ablation, so I offer
this as a design argument, not as a measured control.
\end{faqa}

\cathead{Modeling choices}

\faqq{Your own control shows the EMG-only encoder matching the distilled one on \wer{}. Was the WavLM
distillation therefore wasted?}
\label{faq:distill}
\begin{faqa}
At the word level, effectively yes: 26.10 against 26.14 is a coin flip, and I chose to print that control in
the same table as the result rather than relegate it to a footnote. Two things survive. The distillation does
improve \per{} (22.34 against 23.71), so the encoder is phonetically better even though the decoder cannot
convert the improvement into words, which is itself a preview of the ceiling diagnosed in
Section~\ref{sec:limit}. More importantly, the negative result is load-bearing: it demonstrates that the
parallel audio is not a hidden crutch supporting the headline number, which is exactly what one needs to know
before extending any of this to silent speech, where audio will not exist. A control that invalidates a
favored component is worth more than the component was.
\end{faqa}

\faqq{Why retain the frozen WavLM-to-phoneme recognizer? Are $L_2$ and \glref{infonce}{InfoNCE} not
sufficient to match the teacher's features?}
\begin{faqa}
$L_2$ and InfoNCE make the projection \emph{close} to WavLM. Close is not the same as useful. A projection
can be smooth, correctly scaled, and well correlated frame-by-frame with the teacher, and still be
phonetically empty: regression toward a high-dimensional target will happily settle on a blurred average that
satisfies the loss while discarding the distinctions that separate one phoneme from the next. The
\glref{frozen}{frozen} recognizer changes what ``close'' is permitted to mean. It is a small network that
already reads phonemes out of real audio, and its weights never move, so pushing the EMG projection through it
and demanding a low CTC loss forces the projection to be \emph{decodable into the correct phonemes} rather
than merely audio-shaped. It is a guard against smooth-but-empty blur.
\end{faqa}

\cathead{The language-model reranker}

\faqq{A 7B language model on a single-subject corpus invites memorization. How do you exclude it?}
\begin{faqa}
By assuming it would occur, and building three controls before trusting the number. First, the reranker sees
only the training split, and its training candidates are produced by two-fold
\glref{crossdecode}{cross-decoding}, each half decoded by a model that never trained on it, so it learns to
repair realistic errors rather than the memorized $\sim$1\% \wer{} the acoustic model achieves on its own
training data. Second, six of the 400 test sentences duplicate training text exactly, so the score is reported
both with and without them (18.53 against 18.75) and the reader may take whichever they find credible. Third,
the \glref{verbatim}{verbatim-recall audit} counts how often free generation emits a reference sentence
\emph{absent} from the candidate list, that is, recalled from memory rather than selected from evidence. The
count is zero.
\end{faqa}

\faqq{The free reranker may write anything it likes. Is 18.53\% simply fluent hallucination?}
\begin{faqa}
That is the failure mode the verbatim-recall audit was constructed to detect, and it returns zero: the free
variant never produces a correct reference that the acoustics did not propose. Its gains come from repairing
candidates that were close but wrong, not from inventing candidates that were absent. The structural argument
is the one that also bounds the system (Section~\ref{sec:limit}). A model hallucinating its way to the answer
would be closing the nine-point gap to the \glref{oracle}{oracle}. It is not. It halts precisely where the
acoustic evidence halts, which is what a well-behaved reranker does and what a hallucinating one would not.
\end{faqa}

\cathead{The acoustic ceiling}

\faqq{Distillation bought 1.4 points of \per{} and no \wer{} at all. The \glref{ensemble}{ensemble} bought
1.4 points of \per{} and 2.7 \wer{}. Why the difference?}
\label{faq:ensemble}
\begin{faqa}
Because \per{} and \wer{} interrogate different objects, and the ensemble moves both. \per{} is a
\glref{greedy}{greedy} argmax over frames: it asks only whether the top symbol is correct. The
\glref{wfst}{WFST} decoder never takes the top symbol; it searches over \emph{paths}, weighing whole
sequences against the \glref{lexicon}{lexicon} and the \glref{lm}{language model}. Averaging two encoders'
log-probabilities sharpens the argmax, which is the 1.4 points of \per{}, but it also reshapes the tails of
the distribution, and the tails are where the search actually operates. Paths on which the two models
disagree get suppressed; correct paths that were never top-ranked survive to be found.

I did not separate the two contributions, and this report should not claim to have. The distillation control
offers a rough calibration: there, 1.37 points of \per{} (23.71 to 22.34) bought $-0.04$ \wer{}. If that
exchange rate held at 21--22\% \per{}, the ensemble's 1.44 points would likewise buy nothing and the 2.67
\wer{} points would be the reshaped posterior alone. But that assumes the marginal \wer{} value of a \per{}
point is constant across the range, which it probably is not: a word needs all of its phonemes, so the curve
is likely convex and the lower-\per{} points worth more. Some of the 2.67 may be phonetic after all.

Either answer leaves Section~\ref{sec:limit} standing, which is why the question can be left open here. If the
gain is phonetic, it is direct evidence that \per{} is the quantity that binds. If it is diversity, the
ensemble redistributes information rather than adding any. It cannot rescue the system either way: 2.7 \wer{}
points against the 14.49 that full context bought, and a phone error rate still at 20.9\%.
\end{faqa}

\faqq{Is 20.9\% \per{} a ceiling for EMG, or only for \emph{your} EMG model?}
\begin{faqa}
Only for mine. A single-subject study cannot support the stronger claim. What the report does establish is
narrower and, I think, more useful: for this acoustic model on this corpus, 20.9\% \per{} is the binding
constraint, and no amount of decoder or language-model sophistication passes it. Section~\ref{sec:limit} gives
three independent lines of evidence that the ceiling is acoustic. \per{} moves only in small, expensive
increments under interventions that move \wer{}; the audio teacher cannot transfer its own phonetic quality;
and the nine-point oracle gap consists of words the posteriors never proposed. Whether the ceiling is a
property of surface EMG itself, of the covariance front-end, or of possessing 8{,}500 sentences from one
person, this study cannot distinguish. That is the next experiment. MONA LISA's 3.7\% on a different corpus
(\ref{faq:mona}) is direct evidence that it is not a property of surface EMG itself.
\end{faqa}

\faqq{Is there any independent support for the claim that the reranker, not the encoder, is the exhausted
component?}
\begin{faqa}
Yes, and it arrives from a different signal entirely, which is the best kind of corroboration. Brain2Qwerty
v2~\citep{zhang2026brain2qwerty2} decodes typed sentences from non-invasive MEG with a
\glref{conformer}{Conformer} \glref{ctc}{CTC} encoder and a \glref{lora}{LoRA}-fine-tuned LLM above it, which
is structurally the same stack as Sections~\ref{sec:acoustic} and~\ref{sec:final}. Their encoder alone reaches
55\% \wer{}; an $n$-gram language model takes it to 43\%; the fine-tuned 4B LLM takes it to 39\%. The LLM is
worth four points over the $n$-gram, and it buys them while \emph{worsening} the character error rate: it
trades local accuracy for fluency. The analogy is in the conclusion rather than the mechanism, and the
difference matters. Their LLM rewrites the sentence autoregressively and reads MEG-derived word embeddings
directly, so it can degrade \cer{}; my reranker operates on words downstream of the greedy CTC output and
cannot move \per{} at all, which is why \per{} here stays pinned at 20.9\% rather than worsening. Their own
text-only ablation, which leaves the LLM nothing but the encoder's characters, is the closer analogue to what
I built, and it is worth about six \wer{} points. Their conclusion is nonetheless mine: final performance is
dominated by upstream encoder quality, and improving the encoder is the priority. Two systems, two signals,
one wall.
\end{faqa}

\faqq{What is the single highest-value next step toward sub-10\% \wer{}?}
\begin{faqa}
Not the decoder. Everything downstream of the encoder is exhausted, which is the finding, and the reason
Section~\ref{sec:limit} was written. The gain has to come from the acoustic model: more data, multiple
subjects, or a front-end that recovers phonemes from the raw 31-channel signal rather than from a 25\,ms
covariance summary. If forced to choose one, I would choose additional speakers, because it attacks the
ceiling and the generalization limitation at the same time, and because a single-subject corpus is not a
constraint one can model around.
\end{faqa}

\cathead{Reproducibility and deployment}

\faqq{What did this cost to train, and can it be reproduced?}
\begin{faqa}
The code, the configurations, and, critically, the decode settings are open-sourced under the MIT license at
\href{https://github.com/Varshith-0/myovox}{\texttt{github.com/Varshith-0/myovox}}. That includes the
\glref{blank}{blank penalty}, the \glref{ckpt}{checkpoint-selection criterion}, the regenerated
\texttt{words.txt}, and the \glref{scale}{acoustic scale}: the settings whose combined absence is why the
released configuration decodes at roughly 75\% \wer{} rather than the 40.63\% this baseline reaches. The
blank penalty is much the largest of the four; I did not separate them further. The data is the
\emph{emg2speech} General Corpus, and must be obtained from its authors. Training ran on a single NVIDIA RTX
3080 Ti with 12\,GB of VRAM, and took roughly 15 to 25 hours end to end depending on the experiment (baseline
reproduction, Conformer training, or reranker fine-tuning). The \glref{qlora}{QLoRA} configuration was chosen
precisely so that the 7B reranker fits on that same 12\,GB consumer card.
\end{faqa}

\faqq{Does it run in real time?}
\begin{faqa}
No, and this is a deliberate trade rather than an oversight. The central architectural decision of
Section~\ref{sec:acoustic}, replacing the \glref{causal}{causal} encoder with a \glref{bidir}{bidirectional}
Conformer, lets the model attend to the entire utterance, future frames included, so no word can be emitted
until the sentence ends. The \glref{ensemble}{ensemble}, the multi-scale \glref{nbest}{$n$-best union}, and the
7B \glref{rerank}{reranker} above it are all offline by construction. A \glref{streaming}{streaming} variant is
certainly buildable, and the released causal TDS encoder is one, but it would surrender most of the 14.49
\wer{} points that full context purchased. The objective of this report was to locate the ceiling, not to ship
an interface.
\end{faqa}

\section*{Acknowledgments}

Throughout the project I made heavy use of Anthropic's Claude: for working through the decode-time failures,
structuring the distillation objective and the leakage controls, sifting the literature, and drafting this
report. The design decisions, the experiments, and any errors that remain are mine.

\bibliography{references}

@inproceedings{gowda2026emg2speech,
  title     = {emg2speech: Synthesizing Speech from Electromyography using Self-Supervised Speech Models},
  author    = {Gowda, Harshavardhana T. and Comstock, Daniel C. and Miller, Lee M.},
  booktitle = {Proceedings of the 64th Annual Meeting of the Association for Computational Linguistics (ACL)},
  year      = {2026},
  note      = {arXiv:2510.23969. Baseline reproduced here (Appendix D.4, ``emg2text'').}
}

@article{gowda2025geometric,
  title   = {Non-invasive Electromyographic Speech Neuroprosthesis: A Geometric Perspective},
  author  = {Gowda, Harshavardhana T. and Miller, Lee M.},
  journal = {arXiv preprint arXiv:2502.05762},
  year    = {2025}
}

@article{gowda2024geometry,
  title   = {Geometry of Orofacial Neuromuscular Signals: Speech Articulation Decoding using Surface Electromyography},
  author  = {Gowda, Harshavardhana T. and McNaughton, Zachary D. and Miller, Lee M.},
  journal = {Journal of Neural Engineering},
  year    = {2024}
}

@article{li2024dcond,
  title   = {Brain-to-Text Decoding with Context-Aware Neural Representations and Large Language Models},
  author  = {Li, Jingyuan and Le, Trung and Fan, Chaofei and Chen, Mingfei and Shlizerman, Eli},
  journal = {arXiv preprint arXiv:2411.10657},
  year    = {2024},
  note    = {DCoND-LIFT. Journal of Neural Engineering, 2025.}
}

@inproceedings{hannun2019tds,
  title     = {Sequence-to-Sequence Speech Recognition with Time-Depth Separable Convolutions},
  author    = {Hannun, Awni and Lee, Ann and Xu, Qiantong and Collobert, Ronan},
  booktitle = {Interspeech},
  year      = {2019},
  note      = {arXiv:1904.02619}
}

@inproceedings{gulati2020conformer,
  title     = {Conformer: Convolution-augmented Transformer for Speech Recognition},
  author    = {Gulati, Anmol and Qin, James and Chiu, Chung-Cheng and Parmar, Niki and Zhang, Yu and Yu, Jiahui and Han, Wei and Wang, Shibo and Zhang, Zhengdong and Wu, Yonghui and Pang, Ruoming},
  booktitle = {Interspeech},
  year      = {2020},
  note      = {arXiv:2005.08100}
}

@inproceedings{graves2006ctc,
  title     = {Connectionist Temporal Classification: Labelling Unsegmented Sequence Data with Recurrent Neural Networks},
  author    = {Graves, Alex and Fern{\'a}ndez, Santiago and Gomez, Faustino and Schmidhuber, J{\"u}rgen},
  booktitle = {Proceedings of the 23rd International Conference on Machine Learning (ICML)},
  pages     = {369--376},
  year      = {2006}
}

@article{hsu2021hubert,
  title   = {HuBERT: Self-Supervised Speech Representation Learning by Masked Prediction of Hidden Units},
  author  = {Hsu, Wei-Ning and Bolte, Benjamin and Tsai, Yao-Hung Hubert and Lakhotia, Kushal and Salakhutdinov, Ruslan and Mohamed, Abdelrahman},
  journal = {IEEE/ACM Transactions on Audio, Speech, and Language Processing},
  volume  = {29},
  pages   = {3451--3460},
  year    = {2021},
  note    = {arXiv:2106.07447}
}

@article{chen2022wavlm,
  title   = {WavLM: Large-Scale Self-Supervised Pre-Training for Full Stack Speech Processing},
  author  = {Chen, Sanyuan and Wang, Chengyi and Chen, Zhengyang and Wu, Yu and Liu, Shujie and Chen, Zhuo and Li, Jinyu and Kanda, Naoyuki and Yoshioka, Takuya and Xiao, Xiong and Wu, Jian and Zhou, Long and Ren, Shuo and Qian, Yanmin and Qian, Yao and Zeng, Michael and Yu, Xiangzhan and Wei, Furu},
  journal = {IEEE Journal of Selected Topics in Signal Processing},
  volume  = {16},
  number  = {6},
  pages   = {1505--1518},
  year    = {2022},
  note    = {arXiv:2110.13900}
}

@article{oord2018cpc,
  title   = {Representation Learning with Contrastive Predictive Coding},
  author  = {van den Oord, Aaron and Li, Yazhe and Vinyals, Oriol},
  journal = {arXiv preprint arXiv:1807.03748},
  year    = {2018}
}

@inproceedings{panayotov2015librispeech,
  title     = {Librispeech: An ASR Corpus Based on Public Domain Audio Books},
  author    = {Panayotov, Vassil and Chen, Guoguo and Povey, Daniel and Khudanpur, Sanjeev},
  booktitle = {IEEE International Conference on Acoustics, Speech and Signal Processing (ICASSP)},
  pages     = {5206--5210},
  year      = {2015}
}

@misc{k2icefall,
  title        = {k2 and icefall: FSA/FST Algorithms and ASR Recipes},
  author       = {Povey, Daniel and others},
  howpublished = {\url{https://github.com/k2-fsa/k2}, \url{https://github.com/k2-fsa/icefall}},
  year         = {2023}
}

@inproceedings{hu2022lora,
  title     = {LoRA: Low-Rank Adaptation of Large Language Models},
  author    = {Hu, Edward J. and Shen, Yelong and Wallis, Phillip and Allen-Zhu, Zeyuan and Li, Yuanzhi and Wang, Shean and Wang, Lu and Chen, Weizhu},
  booktitle = {International Conference on Learning Representations (ICLR)},
  year      = {2022},
  note      = {arXiv:2106.09685}
}

@inproceedings{dettmers2023qlora,
  title     = {QLoRA: Efficient Finetuning of Quantized LLMs},
  author    = {Dettmers, Tim and Pagnoni, Artidoro and Holtzman, Ari and Zettlemoyer, Luke},
  booktitle = {Advances in Neural Information Processing Systems (NeurIPS)},
  year      = {2023},
  note      = {arXiv:2305.14314}
}

@article{qwen2025,
  title   = {Qwen2.5 Technical Report},
  author  = {{Qwen Team}},
  journal = {arXiv preprint arXiv:2412.15115},
  year    = {2024}
}

@inproceedings{gaddy2020digital,
  title     = {Digital Voicing of Silent Speech},
  author    = {Gaddy, David and Klein, Dan},
  booktitle = {Proceedings of the 2020 Conference on Empirical Methods in Natural
               Language Processing (EMNLP)},
  pages     = {5521--5530},
  year      = {2020},
  eprint    = {2010.02960},
  archivePrefix = {arXiv},
  note      = {Introduces the 8-channel, single-speaker silent/vocalized facial EMG corpus},
}

@inproceedings{gaddy2021improved,
  title     = {An Improved Model for Voicing Silent Speech},
  author    = {Gaddy, David and Klein, Dan},
  booktitle = {Proceedings of the 59th Annual Meeting of the Association for
               Computational Linguistics and the 11th International Joint Conference
               on Natural Language Processing (Volume 2: Short Papers)},
  year      = {2021},
  eprint    = {2106.01933},
  archivePrefix = {arXiv},
}

@phdthesis{gaddy2022thesis,
  title  = {Voicing Silent Speech},
  author = {Gaddy, David},
  school = {University of California, Berkeley},
  year   = {2022},
}

@article{benster2024mona,
  title   = {A Cross-Modal Approach to Silent Speech with {LLM}-Enhanced Recognition},
  author  = {Benster, Tyler and Wilson, Guy and Elisha, Reshef and Willett, Francis R.
             and Druckmann, Shaul},
  journal = {arXiv preprint arXiv:2403.05583},
  year    = {2024},
  eprint  = {2403.05583},
  archivePrefix = {arXiv},
  note    = {MONA LISA},
}

@inproceedings{mohapatra2025llms,
  title     = {Can {LLM}s Understand Unvoiced Speech? Exploring {EMG}-to-Text
               Conversion with {LLM}s},
  author    = {Mohapatra, Payal and Pandey, Akash and Zhang, Xiaoyuan and Zhu, Qi},
  booktitle = {Proceedings of the 63rd Annual Meeting of the Association for
               Computational Linguistics (Volume 2: Short Papers)},
  pages     = {703--712},
  year      = {2025},
  address   = {Vienna, Austria},
  eprint    = {2506.00304},
  archivePrefix = {arXiv},
}

@article{levy2025brain2qwerty,
  title   = {Brain-to-Text Decoding: A Non-invasive Approach via Typing},
  author  = {L{\'e}vy, Jarod and Zhang, Mingfang and Pinet, Svetlana and
             Rapin, J{\'e}r{\'e}my and Banville, Hubert and d'Ascoli, St{\'e}phane
             and King, Jean-R{\'e}mi},
  journal = {Nature Neuroscience},
  year    = {2025},
  note    = {Brain2Qwerty v1. arXiv:2502.17480},
  eprint  = {2502.17480},
  archivePrefix = {arXiv},
}

@article{zhang2026brain2qwerty2,
  title   = {Accurate Decoding of Natural Sentences from Non-Invasive Brain Recordings},
  author  = {Zhang, Mingfang and L{\'e}vy, Jarod and Rommel, Cedric and
             Rapin, J{\'e}r{\'e}my and Bel, Corentin and Bonnaire, Julie and
             Nieto, Daniel and Bourdillon, Pierre and Pinet, Svetlana and
             d'Ascoli, St{\'e}phane and Moreau, Thomas and King, Jean-R{\'e}mi},
  journal = {Meta AI technical report},
  year    = {2026},
  note    = {Brain2Qwerty v2, 29 June 2026.
             \url{https://github.com/facebookresearch/brain2qwerty}},
}

\end{document}